\documentclass[letterpaper, 10 pt, conference]{ieeeconf}  

\IEEEoverridecommandlockouts 

\usepackage{times,amsmath,amsfonts,stmaryrd,multirow,subfigure,epsfig,enumerate}
\usepackage{ifpdf}
\usepackage{url}
\usepackage{flushend}
\usepackage{cite}
\usepackage{array}
\usepackage{color}
\usepackage{pdfsync}
\usepackage{textcomp}
\usepackage{graphicx}
\usepackage{algorithm}
\usepackage{algcompatible}
\usepackage{textcomp}

\IEEEoverridecommandlockouts
\title{\LARGE \bf
Text2Action: Generative Adversarial Synthesis from Language to Action
}
\author{Hyemin Ahn, Timothy Ha, Yunho Choi, Hwiyeon Yoo, and Songhwai Oh%
\thanks{
H. Ahn, T. Ha, Y. Choi, H. Yoo, and S. Oh are with
the Department of Electrical and Computer Engineering and ASRI,
Seoul National University, Seoul, 08826, Korea
(e-mail: \{hyemin.ahn, timothy.ha, yunho.choi, hwiyeon.yoo\}@cpslab.snu.ac.kr,
songhwai@snu.ac.kr).
}
}

\begin{document}

\maketitle

\begin{abstract}
In this paper, we propose a generative model which learns the
relationship between language and human action in order to generate a
human action sequence given a sentence describing human behavior.
The proposed generative model is a generative adversarial network
(GAN), which is based on the sequence to sequence (\textsc{Seq2Seq})
model.
Using the proposed generative network, we can synthesize various actions
for a robot or a virtual agent using a text encoder recurrent neural
network (RNN) and an action decoder RNN.
The proposed generative network is trained from 29,770 pairs of
actions and sentence annotations extracted from MSR-Video-to-Text
(MSR-VTT), a large-scale video dataset.
We demonstrate that the network can generate human-like actions which
can be transferred to a Baxter robot, such that the robot performs an
action based on a provided sentence.
Results show that the proposed generative network correctly models the
relationship between language and action and can generate a diverse
set of actions from the same sentence.
\end{abstract}

\section{Introduction}
\textit{``Any human activity is impregnated with language because it takes places in an environment that is build up through language and as language''}\cite{language_action}.
As such, human behavior is deeply related to the natural language in our lives.
A human has the ability to perform an action corresponding to a given sentence,
and conversely one can verbally understand the behavior of an observed person.
If a robot can also perform actions corresponding to a given language
description, it will make the interaction with robots easier.
%

Finding the link between language and action has been a great interest in machine learning.
There are datasets which provide human whole body motions and corresponding word or sentence annotations \cite{japan_dataset, KIT_dataset}.
In additions, there have been attempts for learning the mapping between language and human action \cite{japan_related, KIT_related}.
In \cite{japan_related}, hidden Markov models (HMMs) \cite{HMM} is
used to encode motion primitives and to associate them with words.
\cite{KIT_related} used a sequence to sequence (\textsc{Seq2Seq})
model \cite{seq2seq} to learn the relationship between the natural
language and the human actions.

In this paper, we choose to use a generative adversarial network (GAN) \cite{GAN},
which is a generative model, consisting of a generator $G$ and a discriminator $D$.
$G$ and $D$ plays a two-player minimax game, such that $G$ tries to
create more realistic data that can fool $D$ and $D$ tries to
differentiate between the data generated by $G$ and the real data.
Based on this adversarial training method, it has been shown that GANs
can synthesize realistic high-dimensional new data, which is difficult
to generate through manually designed features. \cite{DCGAN_ex1, DCGAN_ex2, DCGAN_ex3}.
In addition, it has been proven that a GAN has a unique solution,
in which $G$ captures the distribution of the real data and $D$ does
not distinguish the real data from the data generated from $G$
\cite{GAN}.
Thanks to these features of GANs,
our experiment also shows that GANs can generate more realistic action
than the previous work \cite{KIT_related}.

\begin{figure}
\centering
\includegraphics[width=0.95\linewidth]{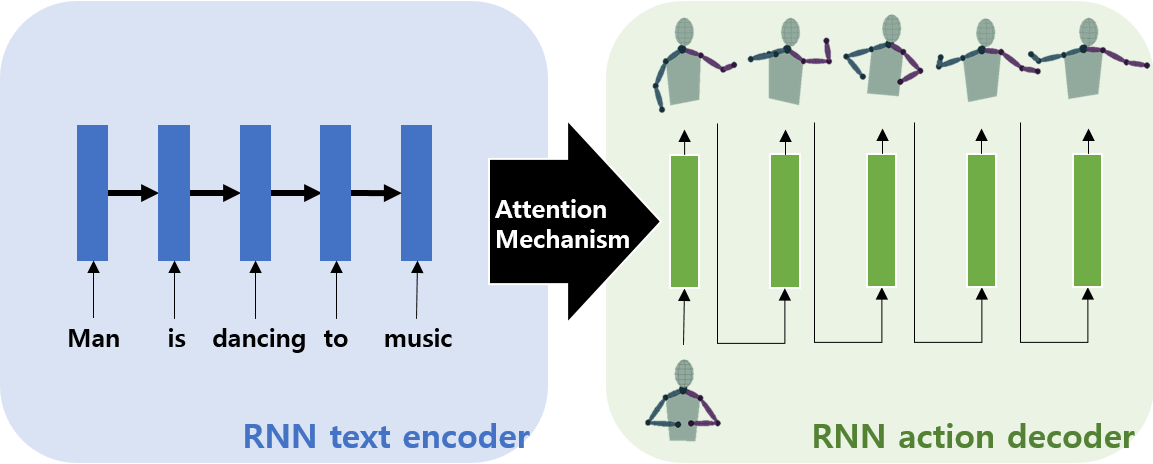}
\caption{
An overview of the proposed generative model.
It is a generative adversarial network \cite{GAN} based on the sequence to sequence model \cite{seq2seq},
which consists of a text encoder and an action decoder based on
recurrent neural networks \cite{RNN}.
When the RNN-based text encoder has processed the input sentence into
a feature vector, the RNN-based action decoder converts the processed
language features to corresponding human actions.
}
\label{fig:overview_example}
\end{figure}

The proposed generative model is a GAN based on the \textsc{Seq2Seq} model.
The objective of a \textsc{Seq2Seq} model is to learn the relationship
between the source sequence and the target sequence, so that it can
generate a sequence in the target domain corresponding to the sequence
in the input domain \cite{seq2seq}.
As shown in Figure~\ref{fig:overview_example},
the proposed model consists of a text encoder and an action decoder
based on recurrent neural networks (RNNs) \cite{RNN}.
Since both sentences and actions are sequences, a RNN is a suitable
model for both the text encoder and action decoder.
The text encoder converts an input sentence, a sequence of words,
into a feature vector.
A set of processed feature vectors is transferred to the action
decoder, where actions corresponding to the input sequence are
generated.
When decoding processed feature vectors,
we have used an attention mechanism based decoder \cite{attention_decoder}.

\begin{figure*}
\centering
\subfigure[]
{
    \centering
    \includegraphics[width=0.48\linewidth]{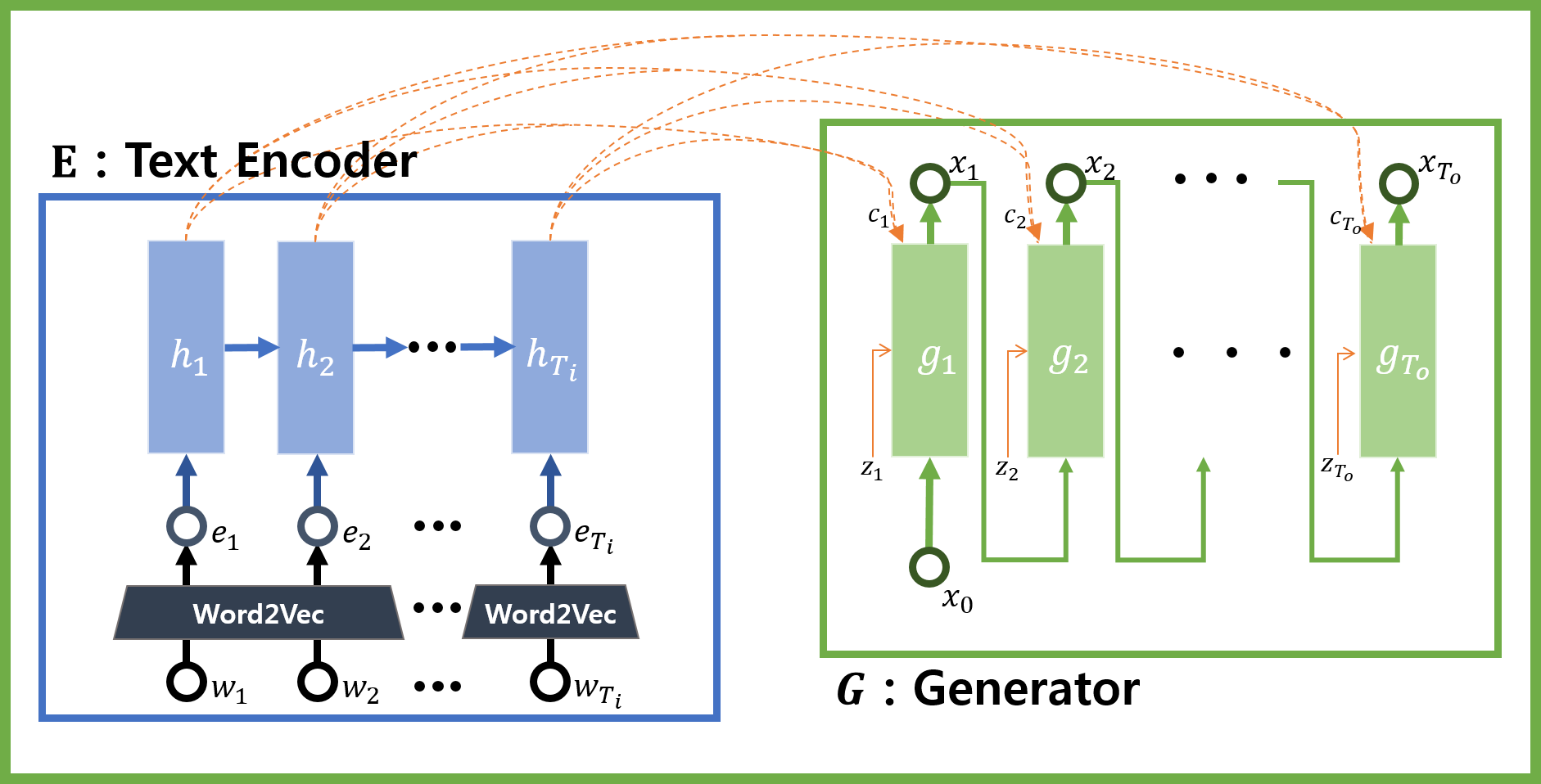}
    \label{fig:generator}
}%
\subfigure[]
{
    \centering
    \includegraphics[width=0.48\linewidth]{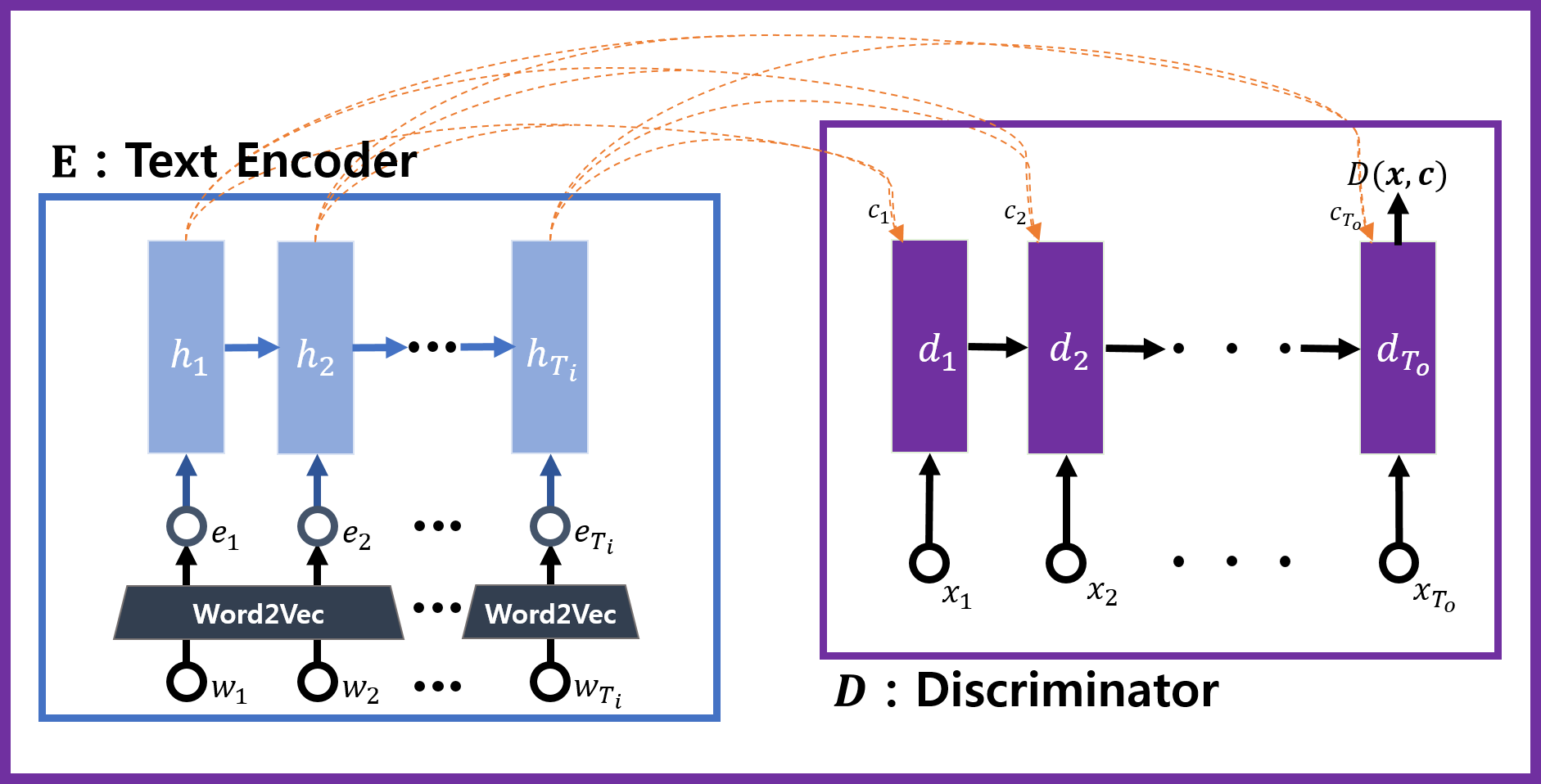}
    \label{fig:discriminator}
}
\caption{
The text encoder $E$, the generator $G$, and the discriminator $D$ constituting the proposed generative model.
The pair of $E$ and $G$, and the pair of $E$ and $D$ are \textsc{Seq2Seq} model composed of the RNN-based encoder and decoder.
Each rectangular denotes the LSTM cell of the RNN.
(a) The text encoder $E$ processes the set of word embedding vectors
$\boldsymbol{e}=\{e_1,\ldots e_{T_i}\}$ into its hidden states
$\boldsymbol{h}=\{h_1,\ldots h_{T_i}\}$.
The generator $G$ takes $\boldsymbol{h}$ and decodes it into the set of feature vectors $\boldsymbol{c}$ and samples the set of random noise vectors $\boldsymbol{z}=\{z_1,\ldots z_{T_o}\}$.
It receives $\boldsymbol{c}$ and $\boldsymbol{z}$ as inputs
and generates the human action sequence $\boldsymbol{x}=\{x_1,\ldots x_{T_o}\}$.
(b)
After the text encoder $E$ encodes the input sentence information into its hidden states $\boldsymbol{h}$,
the discriminator $D$ also takes $\boldsymbol{h}$ and decodes it into the set of feature vectors $\boldsymbol{c}$.
By taking $\boldsymbol{c}$ and $\boldsymbol{x}$ as inputs, $D$
identifies whether $\boldsymbol{x}$ is real or fake.
}
\label{fig:GAN}
\end{figure*}

In order to train the proposed generative network, we have chosen to
use the MSR-Video to Text (MSR-VTT) dataset, which contains web video
clips with annotation texts \cite{MSRVTT}.
Existing datasets \cite{japan_dataset, KIT_dataset} are not suitable
for our purpose since videos are recorded in laboratory environments.
One remaining problem is that the MSR-VTT dataset does not provide
human pose information.
Hence, for each video clip, we have extracted a human upper body pose
sequence using the convolutional pose machine (CPM) \cite{CPM}.
Extracted 2D poses are converted to 3D poses and used as our dataset \cite{2d3d}.
We have gathered $29,770$ pairs of sentence descriptions and action
sequences, containing $3,052$ descriptions and $2,211$ actions.
Each sentence description is paired with about 10 to 12 actions.

The remaining of the paper is constructed as follows.
The proposed Text2Action network is given in Section~\ref{sec:form}.
Section~\ref{sec:method} describes the structure of the proposed generative model
and implementation details.
Section~\ref{sec:exp} shows various 3D human like action
sequences obtained from the proposed generative network and discusses
the result.
In addition, we demonstrate that a Baxter robot can generate the
action based on a provided sentence.

\section{Text2Action Network} \label{sec:form}

Let $\boldsymbol{w}=\{w_1,\ldots w_{T_i}\}$
denote an input sentence composed of $T_i$ words.
Here, $w_t \in \mathbb{R}^d$ is the one-hot vector representation
of the $t\,$th word, where $d$ is the size of vocabulary.
In this paper, we encode $\boldsymbol{w}$ into
$\boldsymbol{e}=\{e_1, \ldots e_{T_i} \}$,
the word embedding vectors for the sentence, based on the word2vec
model \cite{word2vec}.
Here, $e_t \in \mathbb{R}^{n_e}$ is the word embedding representation
of $w_t$, such that $e_t= V w_t$, where $V \in \mathbb{R}^{n_e \times d}$
is the word embedding matrix.
$n_e$ is the dimension of a word embedding vector.
With our dataset, we have pretrained $V$ based on the method presented in \cite{word2vec}.

Since the proposed generative network is a GAN, it consists of
a generator $G$ and a discriminator $D$ as shown in Figure~\ref{fig:GAN}.
The objective of the generator $G$ is to generate a proper human
action sequence corresponding to the input embedding sentence
representation $\boldsymbol{e}$,
and the objective of the discriminator $D$ is to differentiate the
real actions from fake actions considering the given sentence
$\boldsymbol{e}$.
A text encoder, $E$, encodes an embedded sentence, $\boldsymbol{e}$,
into its hidden states, $\boldsymbol{h}=\{h_1, \ldots h_{T_i}\}$,
such that $\boldsymbol{h}$ contains the processed information
related to $\boldsymbol{e}$.
Here, $h_t \in \mathbb{R}^n$ and $n$ is the dimension of the hidden state.

Let $\boldsymbol{x}=\{x_1, \ldots x_{T_o} \}$
denote an action sequence with $T_o$ pose vectors.
Here, $x_t \in \mathbb{R}^{n_x}$ denotes the $t\,$th human pose vector
and $n_x$ is the dimension of a human pose vector.
The pair of the text encoder $E$ and the generator $G$ is a \textsc{Seq2Seq} model.
The generator $G$ converts $\boldsymbol{e}$
into the target human pose sequence $\boldsymbol{x}$.
In order to generate $\boldsymbol{x}$, the generator $G$
decodes the hidden states $\boldsymbol{h}$ of $E$ into a set of
language feature vectors $\boldsymbol{c}=\{c_1 \ldots c_{T_o}\}$ based
on the attention mechanism \cite{attention_decoder}.
Here, $c_t \in \mathbb{R}^{n}$ denotes a feature vector for
generating the $t\,$th human pose $x_t$ and $n$ is the dimension of
the feature vector $c_t$, which is the same as the dimension of
$h_t$.

In addition, a set of random noise vectors
$\boldsymbol{z}=\{z_1,\ldots z_{T_o}\}$ is provided to $G$, where
$z_t \in \mathbb{R}^{n_z}$ is a random noise vector from the zero-mean
Gaussian distribution with a unit variance and
$n_z$ is the dimension of a random noise vector.
With a set of feature vectors $\boldsymbol{c}$ and a set of random
noise vectors $\boldsymbol{z}$, the generator $G$ synthesizes a
corresponding human action sequence $\boldsymbol{x}$, such that
$G(\boldsymbol{z}, \boldsymbol{c}) = \boldsymbol{x}$
(see Figure~\ref{fig:generator}).
Here, the first human pose input $x_0$ is set to the mean pose of all
first human poses in the training dataset.


The objective of the discriminator is to differentiate the $\boldsymbol{x}$ generated from $G$ and the real human action data $\boldsymbol{x}$.
As shown in Figure~\ref{fig:discriminator},
it also decodes the hidden state $\boldsymbol{h}$ of $E$ into the set of language feature vectors $\boldsymbol{c}$ based on the attention mechanism \cite{attention_decoder}.
With a set of feature vectors $\boldsymbol{c}$ and a human action sequence $\boldsymbol{x}$ as inputs,
the discriminator $D$ determines whether $\boldsymbol{x}$ is fake or real considering $\boldsymbol{c}$.
The output from the last RNN cell is the result of the discriminator
such that $D(\boldsymbol{x}, \boldsymbol{c}) \in [0, 1]$
(see Figure~\ref{fig:discriminator}).
The discriminator returns $1$ if the $\boldsymbol{x}$ is identified as real.

In order to train $G$ and $D$, we use the value function defined as follows \cite{GAN}:
\begin{align} \label{eqn:gan_loss}
\min_{G}\max_{D}V(D, G) =&
\mathbb{E}_{\boldsymbol{x}\sim p_{data}(\boldsymbol{x})}
[\log D(\boldsymbol{x}, \boldsymbol{c})]
\\
&+
\mathbb{E}_{\boldsymbol{z}\sim p_{\boldsymbol{z}}(\boldsymbol{z})}
[\log(1-D(G(\boldsymbol{z}, \boldsymbol{c})))]
\nonumber
\end{align}

$G$ and $D$ play a two-player minimax game on the value function $V(D,G)$,
such that $G$ tries to create more realistic data that can fool $D$
and $D$ tries to differentiate between the data generated by $G$ and the real data.

\section{Network Structure} \label{sec:method}

\subsection{RNN-based Text Encoder} \label{sec:encoder}

The RNN-based text encoder $E$ shown in Figure~\ref{fig:GAN}
encodes the input information $\boldsymbol{e}$ into its hidden states of the LSTM cell \cite{RNN}.
Let us denote the hidden states of the text encoder $E$
as $\boldsymbol{h}=\{h_1, \ldots , h_{T_i}\}$,
where
\begin{equation} \label{eqn:q_t}
h_t=q_t(e_t, h_{t-1})\in \mathbb{R}^{n}.
\end{equation}
Here, $n$ is the dimension of the hidden state $h_t$,
and $q_t$ is the nonlinear function in a LSTM cell operating as follows:
\begin{eqnarray}
h_t &=& q_t[e_t, h_{t-1}] = o_t \circ \sigma[C_t]
\\
e'_t &=& W_{e'} e_t + b_{e'}
\label{eqn:h_t}\\
o_t &=& \sigma[W_o e'_t + U_o h_{t-1} + b_o]
\label{eqn:o_t}\\
C_t &=& f_t \circ C_{t-1} + i_t \circ \sigma[W_c e'_t + U_c h_{t-1} + b_c]
\label{eqn:C_t}\\
f_t &=& \sigma[W_f e'_t + U_f h_{t-1} + b_f]
\label{eqn:f_t}\\
i_t &=& \sigma[W_i e'_t + U_i h_{t-1} + b_i]
\label{eqn:i_t}
\end{eqnarray}
where $\circ$ denotes the element-wise production and the $\sigma[x]=\frac{1}{1+e^{-x}}$ denotes the sigmoid function.
The dimension of the matrices and vectors are as follows:
$W_o$, $W_c$, $W_f$, $W_i$, $U_o$, $U_c$, $U_f$, $U_i$ $\in \mathbb{R}^{n \times n}$,
$W_{e'} \in \mathbb{R}^{n \times n_e}$,
$b_o$, $b_c$, $b_f$, $b_i$ $\in \mathbb{R}^n$,
and
$b_{e'} \in \mathbb{R}^{n_e}$.

\subsection{Generator} \label{sec:t2a_gen}

After the text encoder $E$ encodes $\boldsymbol{e}$ into its hidden states $\boldsymbol{h}$,
the generator $G$ decodes $\boldsymbol{h}$ into
the set of feature vectors $\boldsymbol{c}=\{c_1, \ldots c_{T_o}\}$
based on the attention mechanism \cite{attention_decoder},
where $c_t \in \mathbb{R}^n$ is calculated as follows:

\begin{equation} \label{eqn:c_t}
c_t=\sum_{i=1}^{T_o} \alpha_{ti}h_i.
\end{equation}
The weight $\alpha_{ti}$ of each feature $h_i$ is computed as
\begin{equation}
\alpha_{ti}=\frac{\exp(\beta_{ti})}{\sum_{k=1}^{T_o}\exp(\beta_{tk})},
\end{equation}
where
\begin{equation} \label{eqn:beta}
\beta_{ti}=a[g_{t-1}, h_i]=v_a^\top \tanh[W_a g_{t-1} + U_a h_i + b_a].
\end{equation}
Here, the dimensions of matrices and vectors are as follows:
$W_a, U_a \in \mathbb{R}^{n \times n}$,
$v_a, b_a \in \mathbb{R}^n$.

After encoding the language feature $\boldsymbol{c}$,
a set of random noise vectors $\boldsymbol{z}$ is provided to $G$.
With $\boldsymbol{c}$ and $\boldsymbol{z}$, the generator $G$ synthesizes a corresponding human action sequence $\boldsymbol{x}$
such that $G(\boldsymbol{z}, \boldsymbol{c}) = \boldsymbol{x}$.
Let $\boldsymbol{g} = \{ g_1, \ldots g_{T_o} \} $ denote the hidden states of the LSTM cells composing $G$.
Each hidden state of the LSTM cell $g_t \in \mathbb{R}^n$,
where $n$ is the dimension of the hidden state, is computed as follows:
\begin{eqnarray}
g_t
&=& \gamma_t[g_{t-1}, x_{t-1}, c_t, z_t]
\nonumber
\\
&=& W_g (o'_t \circ C'_t) + U_g c_t + H_s z_t + b_g
\label{eqn:g_t}
\\
o'_t &=& \sigma[W_{o'} x'_t + U_{o'} g_{t-1} + b_{o'}]
\\
x'_t &=& W_{x'} x_{t-1} + U_{x'} c_t + H_{x'} z_t + b_{x'}
\\
C'_t &=& f'_t \circ C'_{t-1} + i'_t \circ \sigma[W_{c'} x'_t + U_{c'} g_{t-1} + b_{c'}]
\\
f'_t &=& \sigma[W_{f'}x'_t + U_{f'}g_{t-1} + b_{f'}]
\\
i'_t &=& \sigma[W_{i'}x'_t + U_{i'}g_{t-1} + b_{i'}]
\label{eqn:i'_t}
\end{eqnarray}

and the output pose at time $t$, is computed as
\begin{equation} \label{eqn:x_t}
x_t = W_x g_t + b_x
\end{equation}

The dimensions of matrices and vectors are as follows:
$W_g$, $W_{o'}$, $W_{c'}$, $W_{f'}$, $W_{i'}$, $U_g$, $U_{o'}$, $U_{x'}$, $U_{c'}$, $U_{f'}$, $U_{i'}$ $\in \mathbb{R}^{n \times n}$,
$W_{x'} \in \mathbb{R}^{n \times n_x}$,
$W_x \in \mathbb{R}^{n_x \times n}$,
$H_s, H_{x'} \in \mathbb{R}^{n \times n_z}$,
$b_g, b_{o'}, b_{x'}, b_{c'}, b_{f'}, b_{i'} \in \mathbb{R}^{n}$,
and $b_x \in \mathbb{R}^{n_x}$.
$\gamma_t$ is the nonlinear function constructed based on the attention mechanism presented in \cite{attention_decoder}.

\subsection{Discriminator} \label{sec:t2a_dis}

The discriminator $D$ also decodes $\boldsymbol{h}$ into the set of feature vectors $\boldsymbol{c}$ based on the attention mechanism
(see equations (\ref{eqn:c_t})-(\ref{eqn:beta})) \cite{attention_decoder}.
The discriminator $D$ takes $\boldsymbol{c}$ and $\boldsymbol{x}$ as inputs and generates its scalar value result such that $D(\boldsymbol{x}, \boldsymbol{c}) \in [0, 1]$ (see Figure~\ref{fig:discriminator}).
It returns 1 if the input $\boldsymbol{x}$ has been determined as the real data.
Let
$\boldsymbol{d}=\{d_1, \ldots d_{T_o}\}$ denote the hidden states of the LSTM cell composing $D$,
where $d_t \in \mathbb{R}^{n}$ and $n$ is the dimension of the hidden state $d_t$ which is same as the one of $g_t$.
The output of $D$ is calculated from its last hidden state as follows:
\begin{equation}
D(\boldsymbol{x}, \boldsymbol{c})=\sigma[W_d d_{T_o} + b_d],
\nonumber
\end{equation}
where $W_d \in \mathbb{R}^{1 \times n}$, $b_d \in \mathbb{R}$.
The hidden state of $D$ is computed as
$d_t = \gamma_t [d_{t-1}, x_t, c_t, w_t]$ as similar in (\ref{eqn:g_t})-(\ref{eqn:i'_t}),
where $w_t \in \mathbb{R}^{n_z}$ is the zero vector instead of the random vector $z_t$ such that $w_t =[0, \ldots 0]^\top$.

\begin{figure*}
\centering
\includegraphics[width=0.95\linewidth]{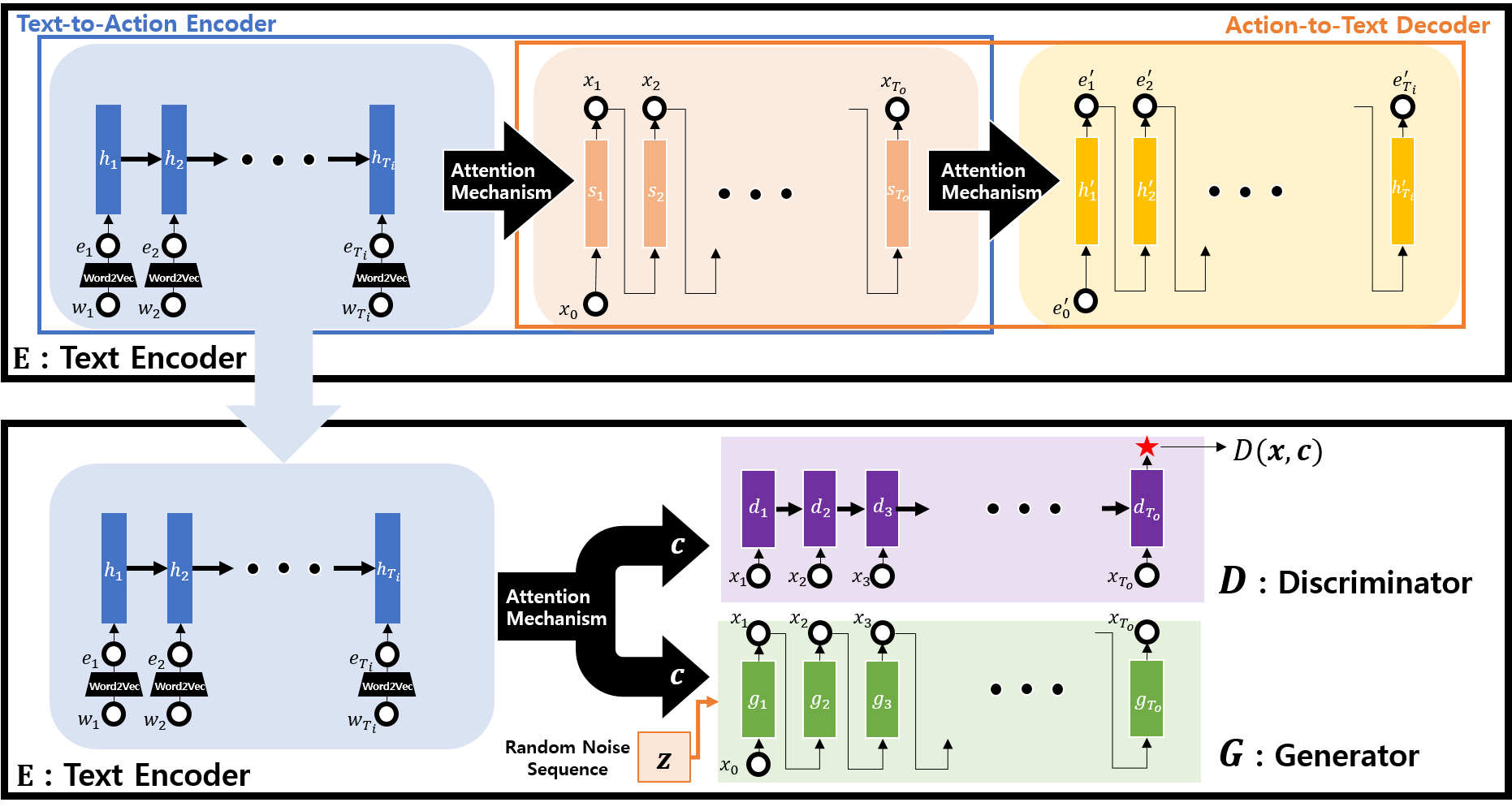}
\caption{
The overall structure of proposed network.
First, we train an autoencoder which maps between the natural language and the human motion.
Its encoder maps the natural language to the human action,
and decoder maps the human action to the natural language.
After training this autoencoder, only the encoder part is extracted
in order to generate the conditional information related to the input sentence so that $G$ and $D$ can use.
}
\label{fig:overall}
\end{figure*}

\subsection{Implementation Details} \label{sec:details}
The desired performance was not obtained properly
when we tried to train the entire network end-to-end.
Therefore, we pretrain the RNN-text encoder $E$ first.
Regarding this, the text encoder $E$ is trained
by training an autoencoder which learns the relationship
between the natural language and the human action as shown in Figure~\ref{fig:overall}.
This autoencoder consists of the text-to-action encoder
that maps the natural language to the human action,
and the action-to-text decoder which reconstructs the human action back to the natural language description.
Both the text-to-action encoder and the action-to-text decoder are \textsc{Seq2Seq} models based on the attention mechanism \cite{attention_decoder}.

The encoding part of the text-to-action encoder corresponds
to the text encoder $E$ in our network,
such that it encodes $\boldsymbol{e}$
into its encoder's hidden states $\boldsymbol{h}$
using (\ref{eqn:q_t})-(\ref{eqn:i_t}).
Based on $\boldsymbol{c}$ (see (\ref{eqn:c_t})-(\ref{eqn:beta})),
the hidden states of its decoder
$\boldsymbol{s} = \{s_1, \ldots s_{T_o}\}$
is calculated as $s_t=\gamma_t[s_{t-1}, x_{t-1}, c_t, w_t]$
(see (\ref{eqn:g_t})-(\ref{eqn:i'_t})),
and $\boldsymbol{x}$ is generated (see (\ref{eqn:x_t})).
Here, $w_t$ is the zero vector instead of random vector
such that $w_t=[0, \ldots 0]^\top \in \mathbb{R}^{n_z}$.

The action-to-text decoder works on the similar principle as above.
After its encoder encodes the human action sequence $\boldsymbol{x}$ into its hidden states
$\boldsymbol{s}=\{s_1, \ldots, s_{T_o}\}$ (see Figure~\ref{fig:overall}),
the decoding part of the action-to-text decoder decodes $\boldsymbol{s}$ into the set of feature vectors
$\boldsymbol{c}'$ (see (\ref{eqn:c_t})-(\ref{eqn:beta})).
Based on $\boldsymbol{c}'$,
the hidden states of its decoder $\boldsymbol{h}'=\{h'_1, \ldots, h'_{T_i}\}$,
is calculated as $h'_t=\gamma_t[h'_{t-1}, e'_{t-1}, c'_t, w_t]$.
(see (\ref{eqn:g_t})-(\ref{eqn:i'_t})).
From the hidden states $\boldsymbol{h}' =\{h'_1, \ldots h'_{T_i}\}$,
the word embedding representation of the sentence $\boldsymbol{e}$ is reconstructed as $\boldsymbol{e}'= \{e'_1, \ldots e'_{T_i}\}$
(see (\ref{eqn:x_t})).
%

In order to train this autoencoder network, we have used a loss function $\mathcal{L}_a$ defined as follows:
\begin{equation} \label{eqn:seq2seq_loss}
\mathcal{L}_a(\boldsymbol{x}, \boldsymbol{e})
=
\frac{a_1}{T_o} \sum_{t=1}^{T_o}\| x_t - \hat{x}_t \|^2_2
+
\frac{a_2}{T_i} \sum_{t=1}^{T_i}\| e_t - e'_t \|^2_2
\end{equation}
where $\hat{x}_t$ denotes the resulted estimation value of $x_t$.
The constants $a_1$ and $a_2$ are used to control
how much the estimation loss of the action sequence $x_1, \ldots x_{T_o}$ and the reconstruction loss of the word embedding vector sequence $e_1, \ldots e_{T_i}$ should be reduced.

\begin{algorithm}[t]
\caption{Training the Text-to-Action GAN}
\label{alg:overall}
\begin{algorithmic}[1]
\STATEx \textbf{Input:}
a set of $N$ input sentences $\{\boldsymbol{e}_1, \ldots \boldsymbol{e}_N \}$ and $N$ output action sequences $\{\boldsymbol{x}_1, \ldots \boldsymbol{x}_N \}$,
a number of training batch steps $S$, batch size $B$.
\STATE Train the autoencoder between the language and action
\STATE Initialize the text encoder $E$ with trained values
\STATE
Initialize weight matrices and bias vectors of $G$ that are shared with the autoencoder to the trained values
\FOR{$s=1 \ldots S$}
\STATE
Randomly sample $\{\boldsymbol{e}_1, \ldots \boldsymbol{e}_B \}$
and $\{\boldsymbol{x}_1, \ldots \boldsymbol{x}_B \}$
\STATE
Sample out the set of random vectors $\{\boldsymbol{z}_1, \ldots \boldsymbol{z}_B \}$
\STATE
Encode sets of feature vectors $\{\boldsymbol{c}_1, \ldots \boldsymbol{c}_B \}$
\STATE
Generate fake data
$\{G(\boldsymbol{z}_1, \boldsymbol{c}_1), \ldots G(\boldsymbol{z}_B, \boldsymbol{c}_B) \}$
\FOR{$b=1 \ldots B$}
\STATE
$y_r(b) \leftarrow D(\boldsymbol{x}_b, \boldsymbol{c}_b)$
\STATE
$y_f(b) \leftarrow D\big(G(\boldsymbol{z}_b, \boldsymbol{c}_b), \boldsymbol{c}_b\big)$
\ENDFOR
\STATE
$\mathcal{V}_D \leftarrow \frac{1}{B}\sum_{b=1}^{B} ( \log y_r(b) + \log (1-y_f(b)))$
\STATE
$D \leftarrow D + \alpha_D \partial \mathcal{V}_D / \partial D $
\STATE
$\mathcal{V}_G \leftarrow \frac{1}{B}\sum_{b=1}^{B}  \log y_f(b) $
\STATE
$G \leftarrow G + \alpha_G \partial \mathcal{V}_G / \partial G $
\ENDFOR
\end{algorithmic}
\end{algorithm}

Overall steps for training the proposed network are presented in Algorithm~\ref{alg:overall}.
After training the autoencoder network,
the part of the text encoder $E$ is extracted and passed to the generator $G$ and discriminator $D$.
In addition, in order to make the training of $G$ more stable,
the weight matrices and bias vectors of $G$ that are shared with the autoencoder,
$W_x$, $W_g$, $W_{o'}$, $W_{x'}$, $W_c$, $W_{f'}$, $W_{i'}$,
$U_g$, $U_{o'}$, $U_{x'}$, $U_c$, $U_{f'}$, $U_{i'}$,
$b_x$, $b_g$, $b_{o'}$, $b_{x'}$, $b_c$, $b_{f'}$, $b_{i'}$,
are initialized to trained values.
When training $G$ and $D$ with the GAN value function shown in (\ref{eqn:gan_loss}),
we do not train the text encoder $E$.
It is to prevent the pretrained encoded language information
from being corrupted while training the network with the GAN value function.

For training the autoencoder network,
we set the number of training epochs as $250$
with batch size $32$.
The dimension of its hidden state in LSTM cell is set to $n=256$.
The Adam optimizer \cite{Adam} is used to minimize the loss function $\mathcal{L}_a$ and the learning rate is set to $5e-5$.
For parameters $a_1$ and $a_2$ in the loss function $\mathcal{L}_a$,
we use $a_1=1$ and $a_2=5$.
The dimension of the hidden state in the LSTM cell composing $G$ and $D$ is set to $n=256$.
The dimension of the random vector $z_t \in \mathbb{R}^{n_z}$ is set to $n_z=16$,
and it is sampled from the Gaussian noise such that $z_t \sim \mathcal{N}(0, 1)^{16}$.
In order to train $G$ and $D$, we set the number of epochs $400$ with batch size $32$.
The Adam optimizer \cite{Adam} is used to maximize the value function $\mathcal{V}_D$ and $\mathcal{V}_G$,
and each learning rate is set to $\alpha_D=2e-6$ and $\alpha_G=2e-6$.
All values of these parameters are chosen empirically.

Regarding training the generator $G$,
we choose to maximize $\log D(G(\boldsymbol{z}, \boldsymbol{c}), \boldsymbol{c})$
rather than minimizing $1-\log D(G(\boldsymbol{z}, \boldsymbol{c}), \boldsymbol{c})$
since it has been shown to be more effective in practice from many cases
\cite{GAN, DCGAN_ex1}.

\section{Experiment} \label{sec:exp}

\begin{figure}
\centering
\includegraphics[width=\linewidth]{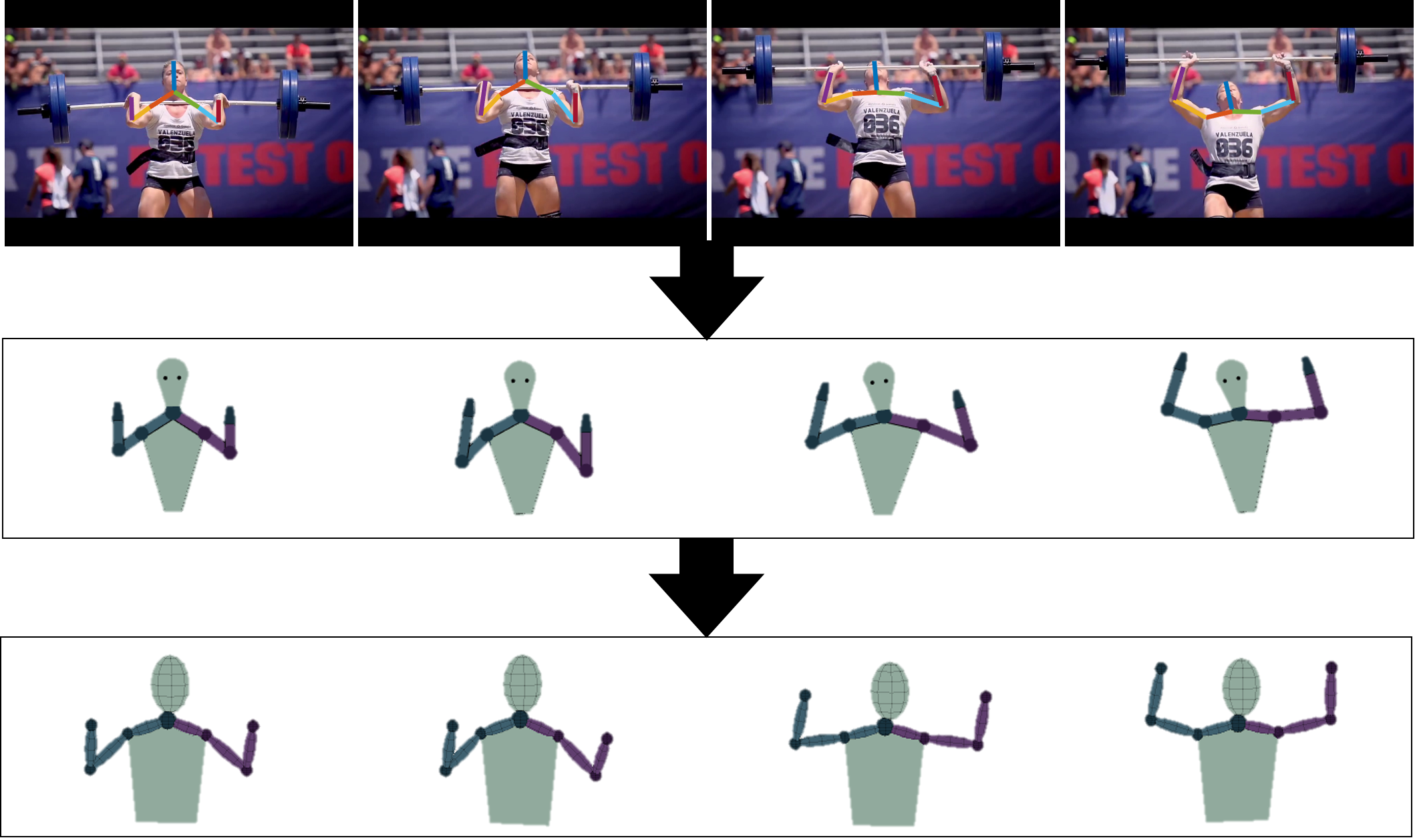}
\caption{
The example dataset for the description `A woman is lifting weights'.
From the video of the MSR-VTT dataset,
we extract the 2D human pose based on the CPM \cite{CPM}.
Extracted 2D human poses are converted to 3D poses based on the code from \cite{2d3d}. The resulting 3D poses are used to train the proposed network.
}
\label{fig:dataset}
\end{figure}

\subsection{Dataset} \label{sec:dataset}

In order to train the proposed generative network,
we use a MSR-VTT dataset which provides Youtube video clips and sentence annotations \cite{MSRVTT}.
As shown in Figure~\ref{fig:dataset},
we have extracted videos in which the human behavior is observed,
and extracted the upper body 2D pose of the observed person through CPM \cite{CPM}.
Extracted 2D poses are converted to 3D poses and used as our dataset \cite{2d3d}.
(The dataset will be made available publicly.)
We choose to use only the upper body pose rather than the full body pose,
since the occlusion near the lower body has been observed in the video considerably.
Another option was to use the data presented in \cite{KIT_dataset},
but there are $6,345$ pairs of actions and sentence description,
which has been judged to be insufficient to train our network.

\begin{figure}
\centering
\includegraphics[width=\linewidth]{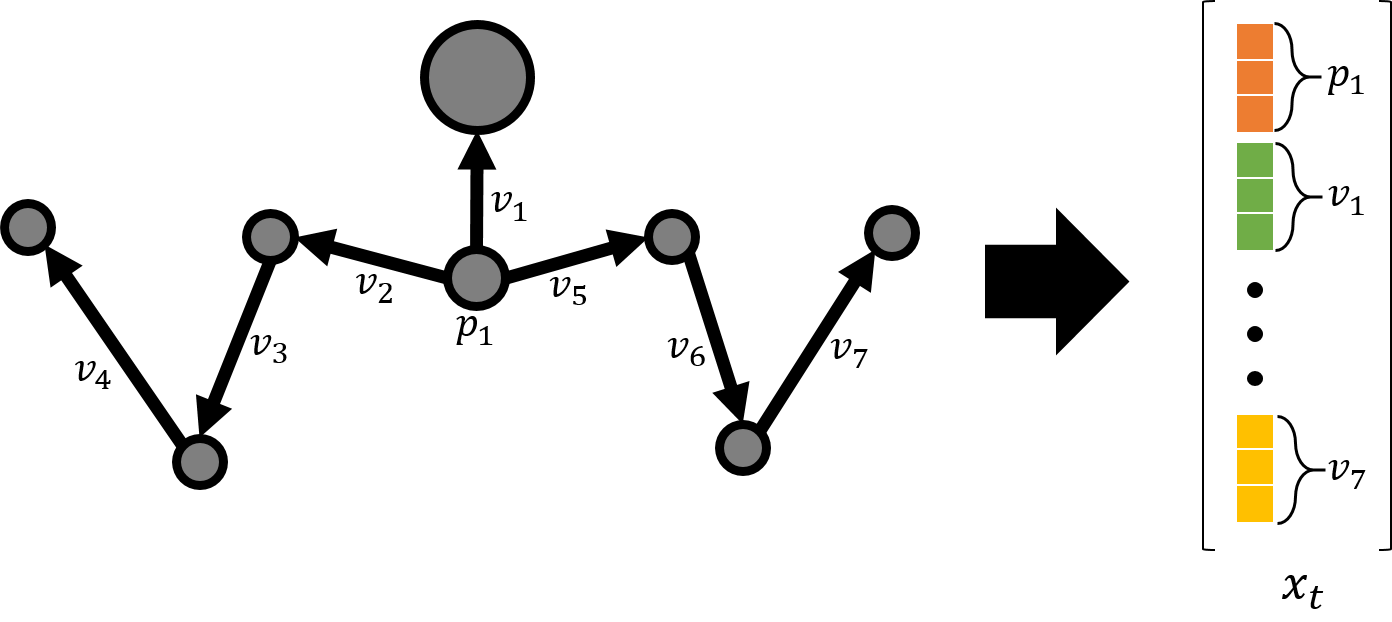}
\caption{
The illustration of how the extracted 3D human pose constructs the
pose vector data $x_t  \in \mathbb{R}^{24}$.
}
\label{fig:pose_data}
\end{figure}

\begin{figure*}[t]
\centering
\includegraphics[width=0.95\linewidth]{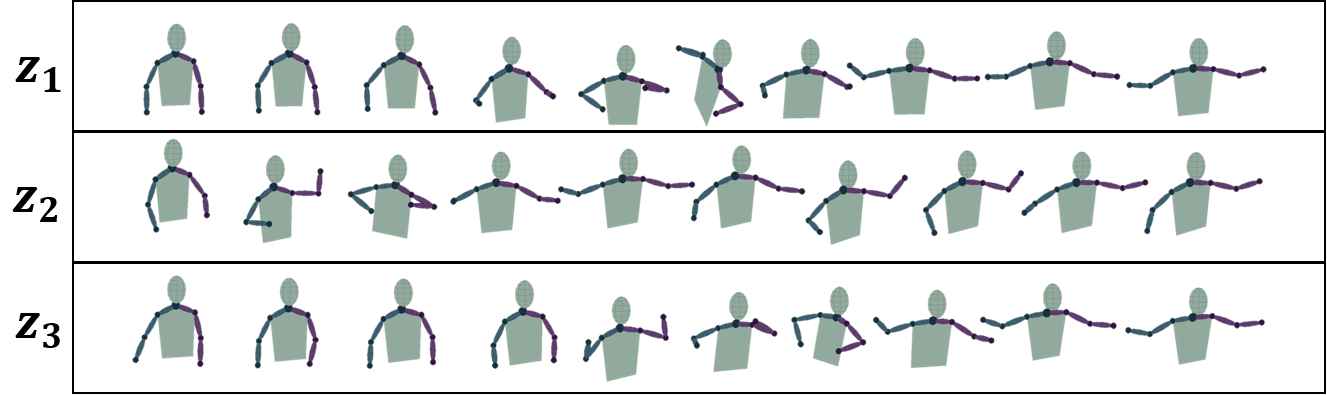}
\caption{
Generated various 3D actions for `A girl is dancing to the hip hop beat'.
$\boldsymbol{z}_1$, $\boldsymbol{z}_2$, and $\boldsymbol{z}_3$ denote the sampled different random noise vector sequences for generating various actions.
}
\label{fig:girl_dance}
\end{figure*}

\begin{figure*}[t]
\centering
\includegraphics[width=0.95\linewidth]{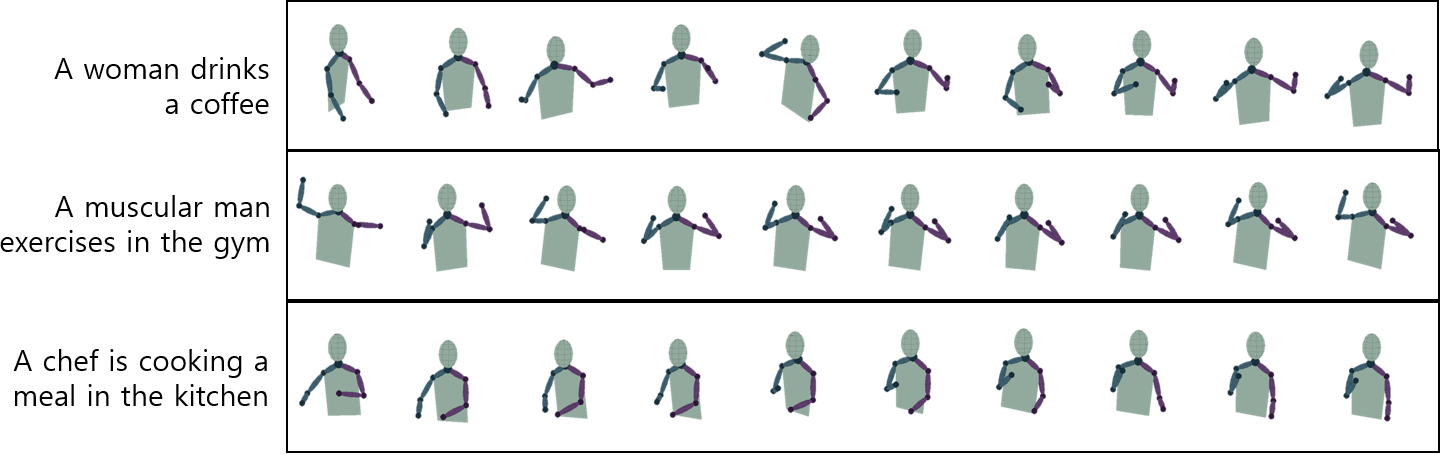}
\caption{
Generated actions when different input sentence is given as input.
When generating these actions, the random noise vector sequence $\boldsymbol{z}$ is fixed and the input feature vectors $\boldsymbol{c}$ are given differently to the generator $G$.
}
\label{fig:various_motions}
\end{figure*}

Each extracted upper body pose for time $t$ is a 24-dimensional vector such that $x_t \in \mathbb{R}^{24}$.
The 3D position of the human neck, and other 3D vectors of seven other joints compose the pose vector data $x_t$ (see Figure~\ref{fig:pose_data}).
Since sizes of detected human poses are different,
we have normalized the joint vectors such that $\|v_i\|_2=1$ for $i=1,\ldots,7$
(see Figure~\ref{fig:pose_data}).
For the poses extracted incorrectly, we manually corrected the pose by hand.
The corrected pose are then smoothed through Gaussian filtering.
Each action sequence is $3.2$ seconds long, and the frame rate is $10$ fps,
making a total of $32$ frames for an action sequence.

Regarding the language annotations,
there were some annotations containing information that is not relevant to the human action.
For example, for a sentence `a man in a brown jacket is addressing the camera while moving his hands wildly',
we cannot know whether the man wears a brown jacket or not
with only human pose information.
For these cases, we manually correct the annotation to include the information only related to the human action
such that `a man is addressing the camera while moving his hands wildly'.

In total, we have gathered $29,770$ pairs of sentence descriptions and action sequences,
which consists of $3,052$ descriptions and $2,211$ actions.
Each sentence description pairs with about 10 to 12 actions.
The time length of total action sequences is $2.713$ hours.
The number of words included in the sentence description data is $21,505$,
and the size of vocabulary which makes up the data is $1,627$.

\subsection{3D Action Generation} \label{sec:3d_result}

We first examine how action sequences are generated
when a fixed sentence input and a different random noise vector inputs are
given to the trained network.
Figure~\ref{fig:girl_dance} shows
three actions generated with one sentence input and three differently sampled random noise vector sequences
such that $G(\boldsymbol{z}_1, \boldsymbol{c}), G(\boldsymbol{z}_2, \boldsymbol{c}), G(\boldsymbol{z}_3, \boldsymbol{c})$.
Generated pose vector data $x_t$ which contains $p_1$, $v_1, \ldots,
v_7$ (see Figure~\ref{fig:pose_data}) is fitted to the human skeleton
of a predetermined size.
The input sentence description is `A girl is dancing to the hip hop beat', which is not included in the training dataset.
In this figure, human poses in a rectangle represent the one action sequence, listed in time order from left to right.
The time interval between the each pose is 0.5 second.
It is interesting to note that
even though the same sentence input is given,
varied human actions are generated if the random vectors are different.
In addition, it is observed that generated motions are all taking the action like dancing.

We also examine how the action sequence is generated
when the input random noise vector sequence $\boldsymbol{z}$ is fixed and the sentence input information $\boldsymbol{c}$ varies.
Figure~\ref{fig:various_motions} shows three actions generated based on the one fixed random noise vector sequence and three different sentence inputs such that $G(\boldsymbol{z}, \boldsymbol{c}_1), G(\boldsymbol{z}, \boldsymbol{c}_2), G(\boldsymbol{z}, \boldsymbol{c}_3)$.
Input sentences are `A woman drinks a coffee',
`A muscular man exercises in the gym', and `A chef is cooking a meal in the kitchen'.
The disadvantage of the given result is that it is difficult to understand the concrete context by only seeing the action, since no tools or background information related to the given action is given.
However,
the first result in Figure~\ref{fig:various_motions} shows the action sequence as if a human is lifting right hand and getting close to the mouth as when drinking something (see the $5\,$th frame).
The second result shows the action sequence like a human moving with a dumbbell in both hands.
The last result shows the action sequence as if a chef is cooking food in the kitchen and trying a sample.

\begin{figure*}
\centering
\includegraphics[width=\linewidth]{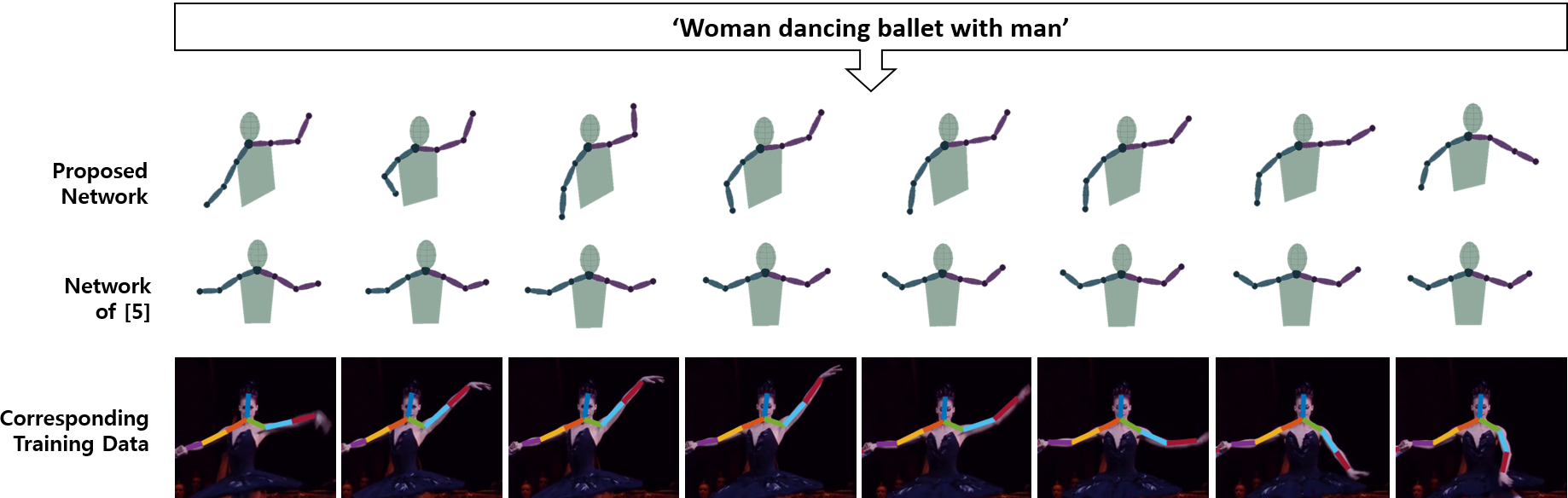}
\caption{
The result of comparison when the input sentence is `Woman dancing ballet with man', which is included in the dataset.
Each result is compared with the human action data that corresponds to the input sentence in the training dataset.
}
\label{fig:compare_ballet}
\end{figure*}

\subsection{Comparison with \cite{KIT_related}} \label{sec:comparison}

In order to see the difference between our proposed network and the network of \cite{KIT_related},
we have implemented the network presented in \cite{KIT_related}
based on the Tensorflow and trained it with our dataset.
First, we compare generated actions when we give the sentence
`Woman dancing ballet with man', which is included in the training dataset,
as an input to the each network.
The result of the comparison is shown in Figure~\ref{fig:compare_ballet}.
The time interval between the each pose is 0.4 second.
In this figure, results from both networks are compared to the human action data that matches to the input sentence in the training dataset.
The result shows that our generative model synthesizes the human action sequence that is more similar to the data.
Although the network presented in \cite{KIT_related} also generates the action as the ballet player with both arms open,
it is shown that the action sequence synthesized by our network is more natural and similar to the data.

\begin{figure*}
\centering
\includegraphics[width=\linewidth]{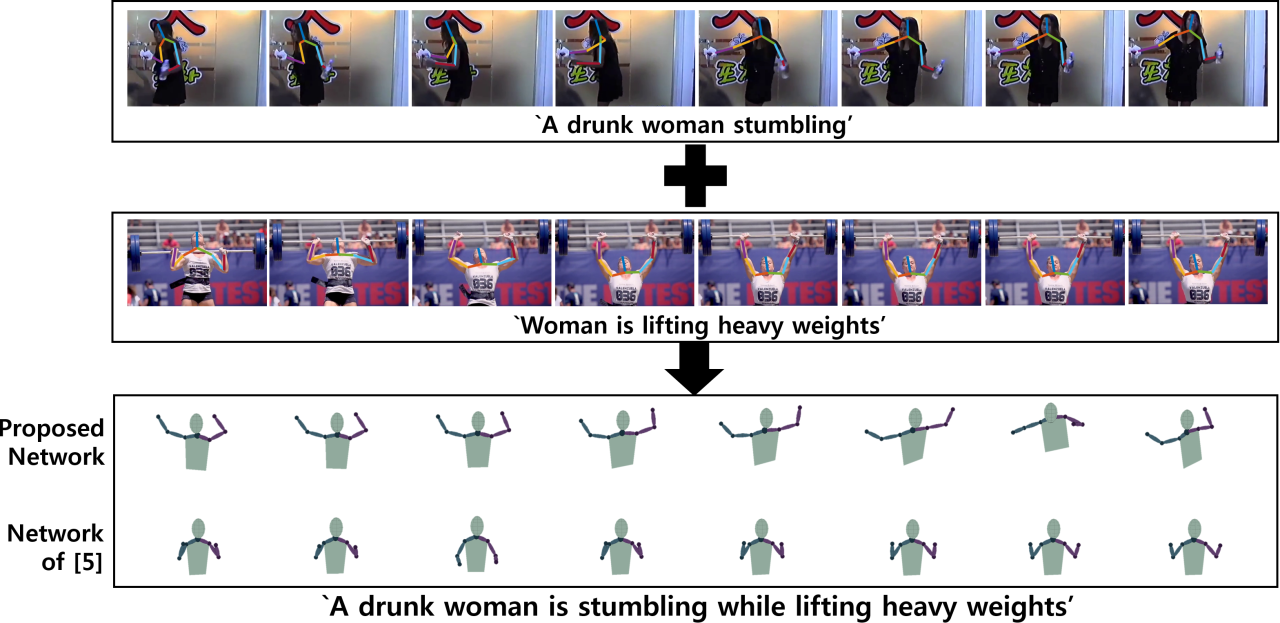}
\caption{
The result of comparison when the sentence which is not included in the dataset is given as an input.
The input sentence given to the each network is `A drunk woman is stumbling while lifting heavy weights'.
This sentence is a combinations of `A drunk woman stumbling' and `Woman is lifting heavy weights', which are included in the training dataset.
The result shows that our proposed network generates the human action sequence
corresponds to the proper combinations of the two training data.
The generated action sequence is like a drunk woman staggering and lifting the weights.
}
\label{fig:compare_stumble}
\end{figure*}

In addition, we give the sentence which is not included in the training dataset
as an input to each network.
The result of the comparison is shown in Figure~\ref{fig:compare_stumble}.
The time interval between the each pose is 0.4 second.
The given sentence is `A drunk woman is stumbling while lifting heavy weights'.
It is a combinations of two sentences included in the training dataset, which are `A drunk woman stumbling' and `Woman is lifting heavy weights'.
Although we know that it is difficult to see the situation as described by the input sentence,
this experiment is to test whether the proposed network has learned well about the relationship between natural language and human action
and responds flexibly to input sentences.
The action sequence generated from our network is like a drunk woman staggering and lifting the weights,
while the action sequence generated from the network in \cite{KIT_related} is just like a person lifting weights.

It is shown that the method suggested in \cite{KIT_related}
also produces the human action sequence that seems somewhat corresponding to the input sentence,
however, the generated behaviors are all symmetric and not as dynamic as the data.
It is because their loss function is designed to maximize the likelihood of the data, whereas the data contains asymmetric pose to the left or right.
As an example of the ballet movement shown in Figure~\ref{fig:compare_ballet},
our training data may have a left arm lifting action and a right arm lifting action to the same sentence `Woman dancing ballet with man'.
But with the network that is trained to maximize the likelihood of the entire data,
a symmetric pose to lift both arms has a higher likelihood and
eventually become a solution of the network.
On the other hand,
our network which is trained based on the GAN value function (\ref{eqn:gan_loss})
manages to generate various human action sequences that look close to the training data.

\begin{figure*}
\centering
\includegraphics[width=0.95\linewidth]{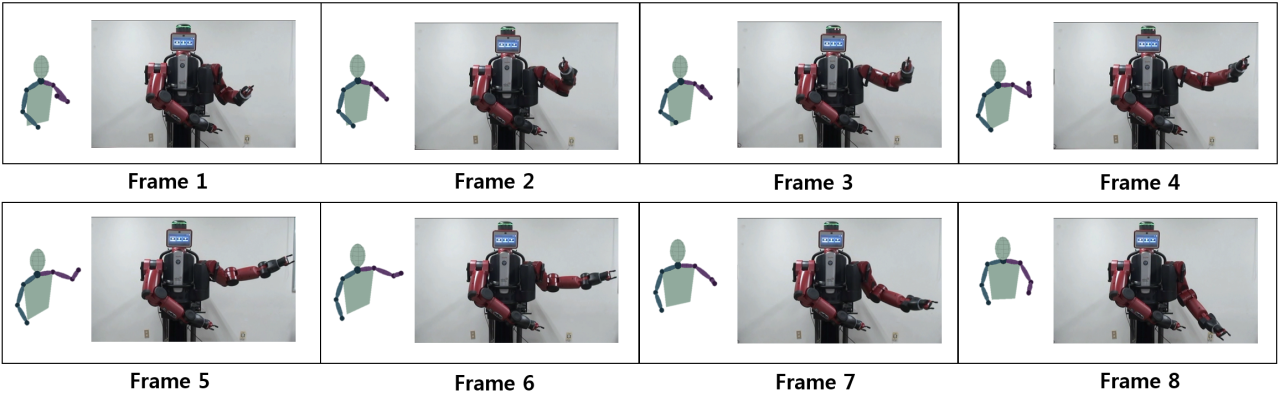}
\caption{
Results of applying generated action sequence to the Baxter robot.
The generated action sequence is applied to the Baxter robot
based on the Baxter-teleoperation code from \cite{baxter_3d}.
The time difference between frames capturing Baxter's pose is about $2$ second.
}
\label{fig:result_baxter}
\end{figure*}

\subsection{Generated Action for a Baxter Robot} \label{sec:baxter}

We enable a Baxter robot to execute the give action trajectory defined
in a 3D Cartesian coordinate system by referring the code from \cite{baxter_3d}.
Since the maximum speed at which a Baxter robot can move its joint is limited, we slow down the given action trajectory and apply it to the robot.
Figure~\ref{fig:result_baxter} shows how the Baxter robot executes the given 3D action trajectory
corresponding to the input sentence `A man is throwing something out'.
Here, the time difference between frames capturing the Baxter's pose is about $2$ second.
We can see that the generated 3D action takes the action as throwing something forward.

\section{Conclusion} \label{sec:con}
In this paper, we have proposed a generative model based on the
\textsc{Seq2Seq} model \cite{seq2seq}
and generative adversarial network (GAN)\cite{GAN},
for enabling a robot to execute various actions corresponding to an
input language description.
In order to train the proposed network,
we have used the MSR-Video to Text dataset \cite{MSRVTT},
which contains recorded videos from real-world situations
and uses a wider range of words in the language description
than other datasets.
Since the data do not contain 3D human pose information,
we have extracted the 2D upper body pose of the observed person
through convolutional pose machine \cite{CPM}.
Extracted 2D poses are converted to 3D poses and used as our dataset \cite{2d3d}.
The generated 3D action sequence is transferred to a robot.

It is interesting to note that our generative model,
which is different from other existing related works
in terms of utilizing the advantages of the GAN,
is able to generate diverse behaviors
when the input random vector sequence changes.
In addition, results show that
our network can generate an action sequence that is more dynamic and
closer to the actual data than the network presented presented in
\cite{KIT_related}.
The proposed generative model, which understands the relationship
between the human language and the action, generates an action
corresponding to the input language.
We believe that the proposed method can make actions by robots more
understandable to their users.

\bibliography{text2action}
\bibliographystyle{IEEEtran}
\end{document}